\documentclass[runningheads]{llncs}
\usepackage[T1]{fontenc}
\usepackage{graphicx}
\usepackage{float}
\usepackage{pifont}
\usepackage{amsmath}
\usepackage{booktabs}
\usepackage{color, colortbl, xcolor, soul}
\usepackage[colorlinks=true,linkcolor=blue, citecolor=blue, urlcolor=blue]{hyperref}

\definecolor{Gray1}{gray}{0.95}
\definecolor{Gray2}{gray}{0.81}

\sethlcolor{yellow}

\newcommand\scalemath[2]{\scalebox{#1}{\mbox{\ensuremath{\displaystyle #2}}}}

\begin{document}
\title{AdaptDiff: Cross-Modality Domain Adaptation via Weak Conditional Semantic Diffusion for Retinal Vessel Segmentation}
\titlerunning{AdaptDiff}

\author{Dewei Hu\inst{1}\and
Hao Li\inst{1} \and
Han Liu\inst{2} \and
Jiacheng Wang\inst{2} \and
Xing Yao\inst{2} \and
Daiwei Lu\inst{2} \and
Ipek Oguz\inst{1}\inst{2}}

\authorrunning{D. Hu et al.}

\institute{Department of Electrical and Computer Engineering, Vanderbilt University \and 
Department of Computer Science, Vanderbilt University\\
\email{hudewei1212@gmail.com}}
\maketitle              
\begin{abstract}
Deep learning has shown remarkable performance in medical image segmentation. However, despite its promise, deep learning has many challenges in practice due to its inability to effectively transition to unseen domains, caused by the inherent data distribution shift and the lack of manual annotations to guide domain adaptation. To tackle this problem, we present an unsupervised domain adaptation (UDA) method named \textit{AdaptDiff} that enables a retinal vessel segmentation network trained on fundus photography (FP) to produce satisfactory results on unseen modalities (e.g., OCT-A) without any manual labels. For all our target domains, we first adopt a segmentation model trained on the source domain to create pseudo-labels. With these pseudo-labels, we train a conditional semantic diffusion probabilistic model to represent the target domain distribution. Experimentally, we show that even with low quality pseudo-labels, the diffusion model can still capture the conditional semantic information. Subsequently, we sample on the target domain with binary vessel masks from the source domain to get paired data, i.e., target domain synthetic images conditioned on the binary vessel map. Finally, we fine-tune the pre-trained segmentation network using the synthetic paired data to mitigate the domain gap. We assess the effectiveness of AdaptDiff on seven publicly available datasets across three distinct modalities. Our results demonstrate a significant improvement in segmentation performance across all unseen datasets. Our code is publicly available at \url{https://github.com/DeweiHu/AdaptDiff}.

\end{abstract}

\section{Introduction}
In recent years, the increasing availability of data and computational resources has propelled the rapid advancement of deep learning techniques in medical image analysis. However, the effectiveness of these techniques hinges on the similarity in distribution between training and testing datasets. This condition is seldom met in medical imaging due to the absence of unified imaging protocols. Consequently, a domain adaptation (DA) \cite{guan2021domain} step is often required for unseen target datasets so that a model trained on source domain can adjust to the distribution shift. Among all the categories of DA, the unsupervised domain adaptation (UDA) is most commonly used as there is often no ground truth available for the unseen target data.  Specifically for the segmentation task,  manual annotation is labor-intensive and time-consuming \cite{kumari2023deep}. The UDA can be particularly challenging when the domain shift is substantial (e.g., in different modalities). 

\begin{figure}[t]
    \centering
    \includegraphics[width=.95\linewidth]{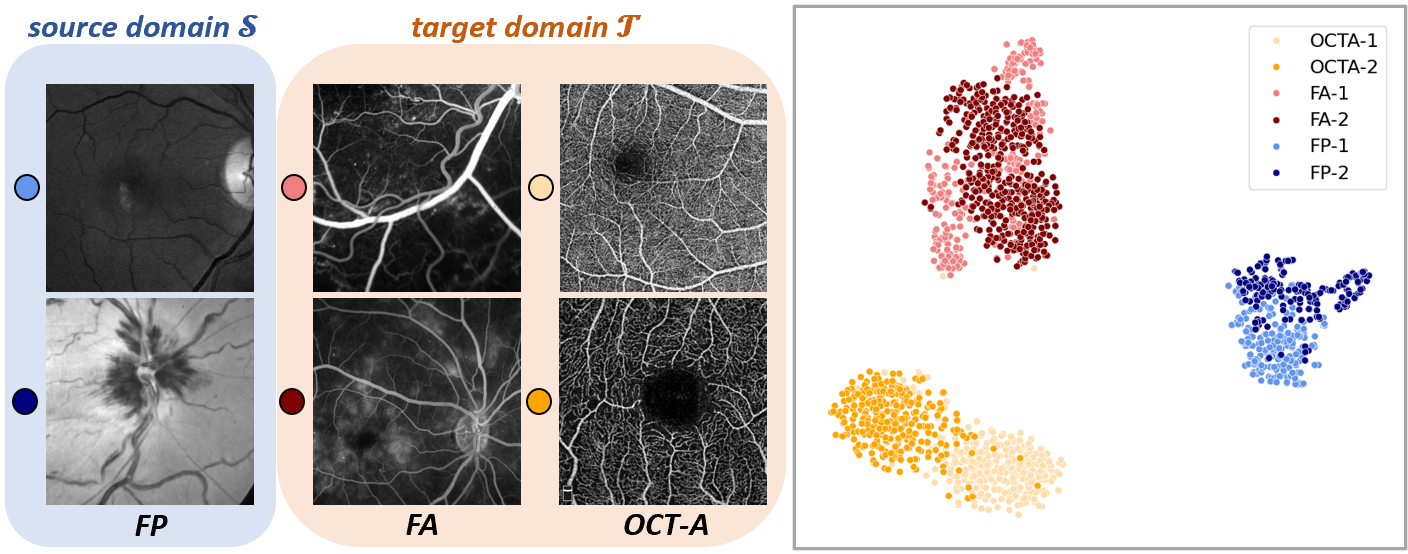}
    \caption{\textbf{Left}: example patches for datasets in FP, FA and OCTA. The outlines color-code the modality (red: FA, yellow: OCT-A, blue: FP). \textbf{Right}: T-SNE plot to visualize the separation between domains in feature space (extracted by pre-trained VGG-16).}
    \label{fig:domains tsne}
\end{figure}

Retinal vasculature serves as a crucial biomarker for various diseases including diabetic retinopathy, glaucoma and age-related macular degeneration \cite{burns2021imaging}. While the standard modalities used to visualize the retinal vessels are OCT angiography (OCT-A) and fluorescein angiography (FA), most manual annotation of retinal vessels are conducted on fundus photography (FP) \cite{budai2013robust,hoover2000locating,farnell2008enhancement,staal2004ridge} as FP is easier to segment than OCT-A, and more widely available than FA.  Availability of labeled public datasets in OCT-A \cite{li2020image,ma2020rose} and FA \cite{ding2020novel} is very limited in contrast. Thus, we propose to use labeled FP datasets as source domains and develop a UDA method that transfers the knowledge to the unlabeled OCT-A and FA target datasets. Fig.~\ref{fig:domains tsne}  provides an example of each dataset and shows the domain shift between the aforementioned modalities. 

Our general idea is to train a conditional diffusion probabilistic model \cite{ho2020denoising} to synthesize target domain image from a given binary vessel mask. Similar models based upon spatially-adaptive normalization block \cite{park2019semantic} has been presented in data augmentation for histology \cite{yu2023diffusion,oh2023diffmix} and colonoscopy \cite{du2023arsdm}. However, all these diffusion models are trained with annotated data in the same modality. In this study, we explore,  for the first time, training the conditional diffusion model in a weakly supervised manner for the cross-modality scenario.


First, using the labeled source domain data, we train a segmentation model. Next, we utilize this model to create pseudo-labels for target domain images. With the pseudo-labeled data, we train a semantic conditional diffusion model  that maps the  vessel masks to the target domain. Experimentally, we show that even with the noisy labels, the resultant diffusion model is still sufficient to represent the target data distribution. \textit{Since both source and target domains are images delineating human retinal vasculature, we assume that they share the same label space.} Therefore, we can generate paired target domain data by sampling with source domain labels. Finally, we fine-tune the segmentation network with the synthetic paired data  to adapt to the target domain. We evaluate AdaptDiff on 7 public datasets (3 FP datasets for training, 2 OCT-A and 2 FA datasets for testing) and show it significantly improves the segmentation performance for cross-modality data. Our main contributions are:
\begin{itemize}
    \item[\ding{117}] We are the first to develop the cross-modality UDA with a conditional diffusion probabilistic model.
    \item[\ding{117}] We are the first to train a conditional diffusion model with pseudo-labels. We experimentally show that the semantic conditional diffusion trained with even very  noisy labels can effectively represent the target distribution. 
    \item[\ding{117}] We conduct extensive experiments on 7 public datasets in 3 distinct modalities to show the superior domain adaptation performance of AdaptDiff. 
\end{itemize}

\section{Related works}
\label{Sec:related work}
We begin with an overview of major previous works about deep UDA, which can be categorized into (1) feature alignment and (2) image alignment. 

\noindent
\normalfont{\textbf{Feature alignment.}} The UDA can be achieved by reducing the disparity between intermediate feature maps extracted from source and target domain images. Such features are also regarded as a domain-invariant representation. In practice, this idea  primarily relies on the mini-max approach in adversarial learning \cite{goodfellow2014generative}. In the domain-adversarial neural network (DANN) \cite{ganin2015unsupervised}, a gradient reversal layer is implemented in an adversarial network framework to enforce the comparability of feature map distributions across domains. Based on DANN, Javanmardi et al.\ \cite{javanmardi2018domain} propose a domain adaptation method on FP vessel segmentation. We use their approach as one of our competing methods.

\noindent
\normalfont{\textbf{Image alignment.}} In contrast, image alignment methods perform domain alignment in image space, by converting  an image in source domain to the style of the target domain. Here,  `style' refers to the appearance characteristics excluding the semantic content. Such unpaired image-to-image style transfer is commonly done using CycleGAN \cite{zhu2017unpaired}. Palladino et al.\ \cite{palladino2020unsupervised} implement this framework in white matter segmentation for multicenter MR images. The contrastive learning for unpaired image translation (CUT) \cite{park2020contrastive} is another method that are often used for image alignment. Huo et al.\ \cite{huo2018synseg} propose an end-to-end framework called SynSeg that combines the cycle adversarial network with the segmentation model. We implement these three methods for comparison. 


\section{Methods}
\label{Sec:method}

\begin{figure}[t]
\centering
\begin{tabular}{cccc}
\includegraphics[width=.24\linewidth]{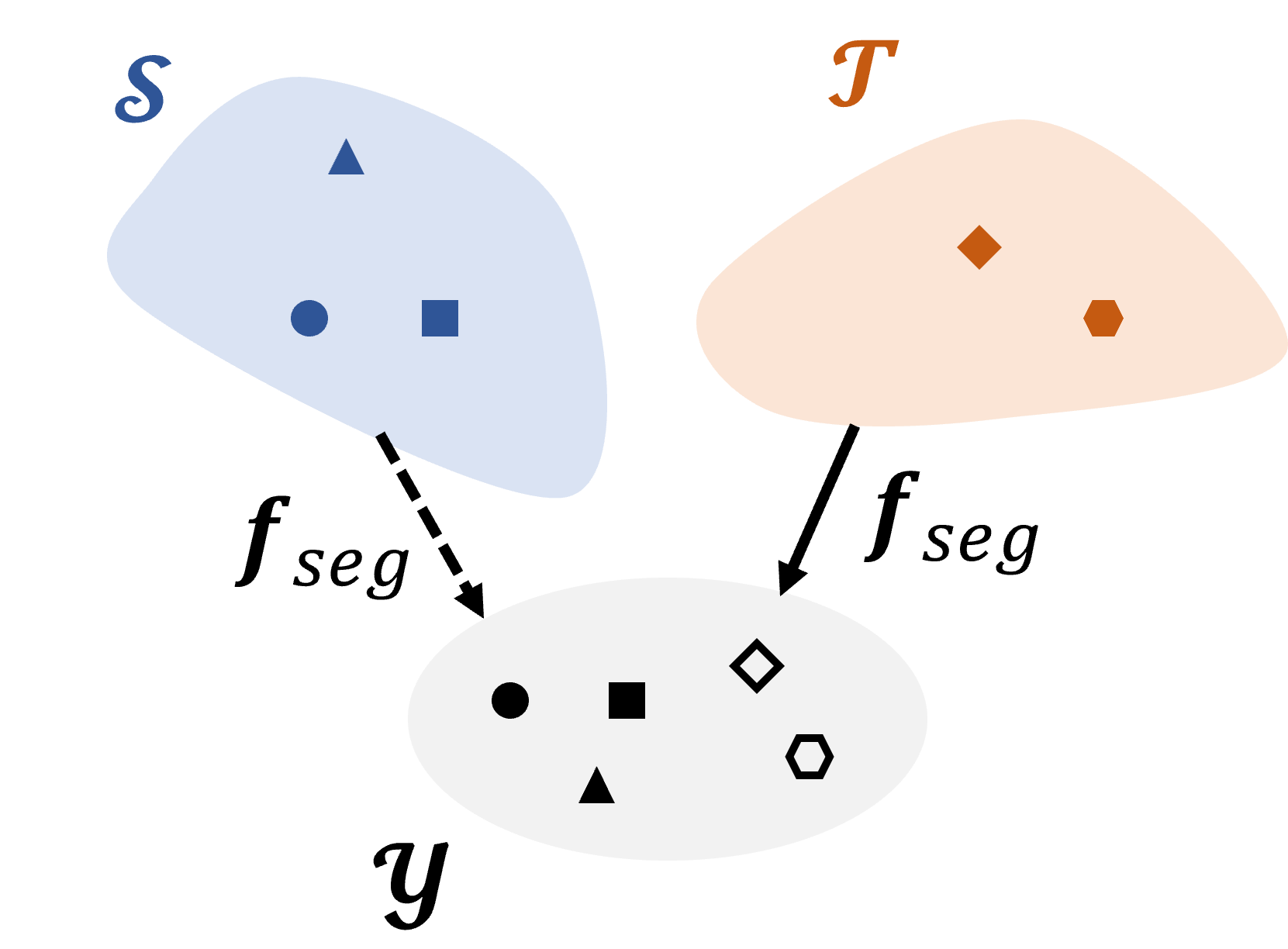}&
\includegraphics[width=.24\linewidth]{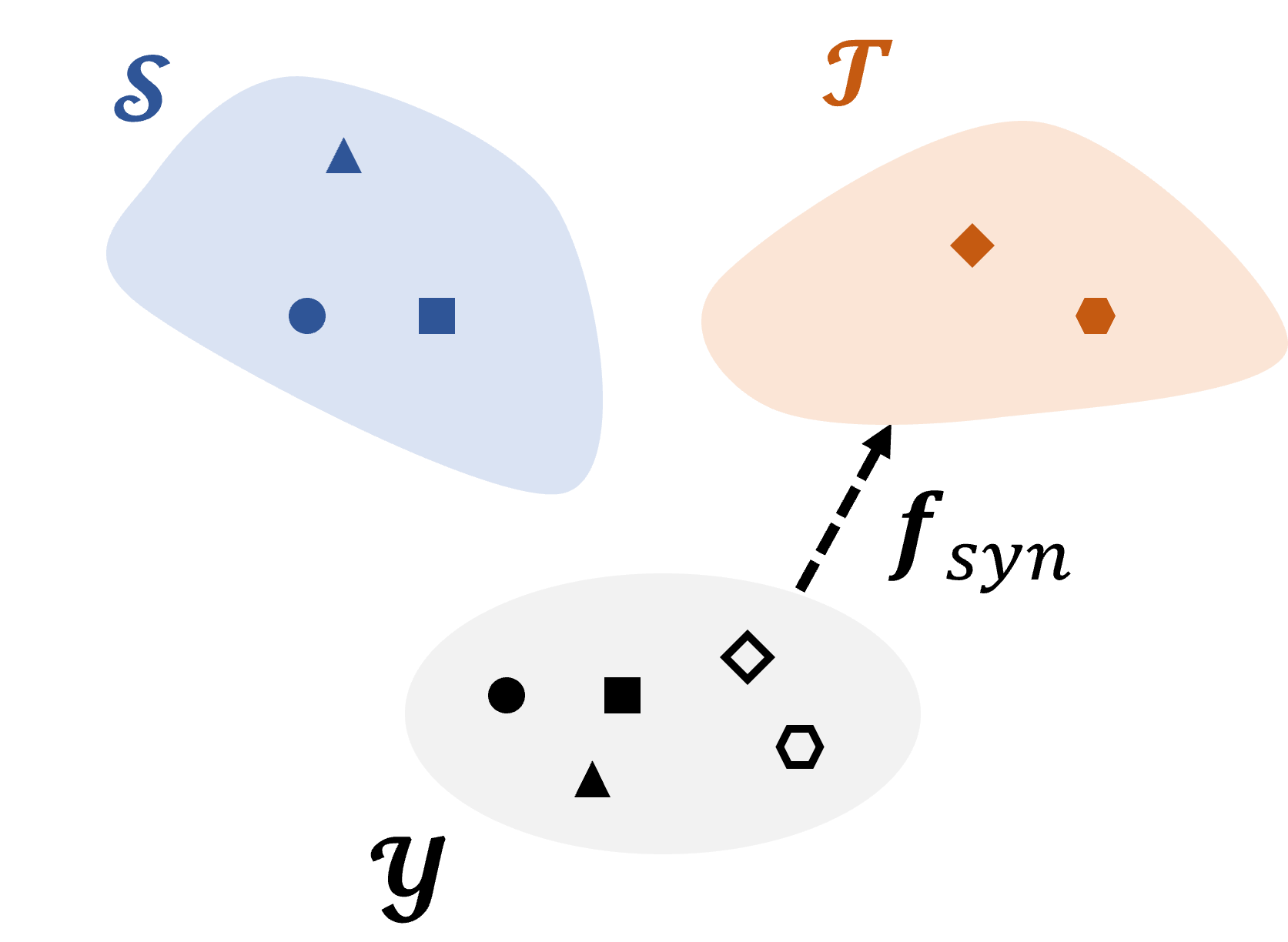}&
\includegraphics[width=.24\linewidth]{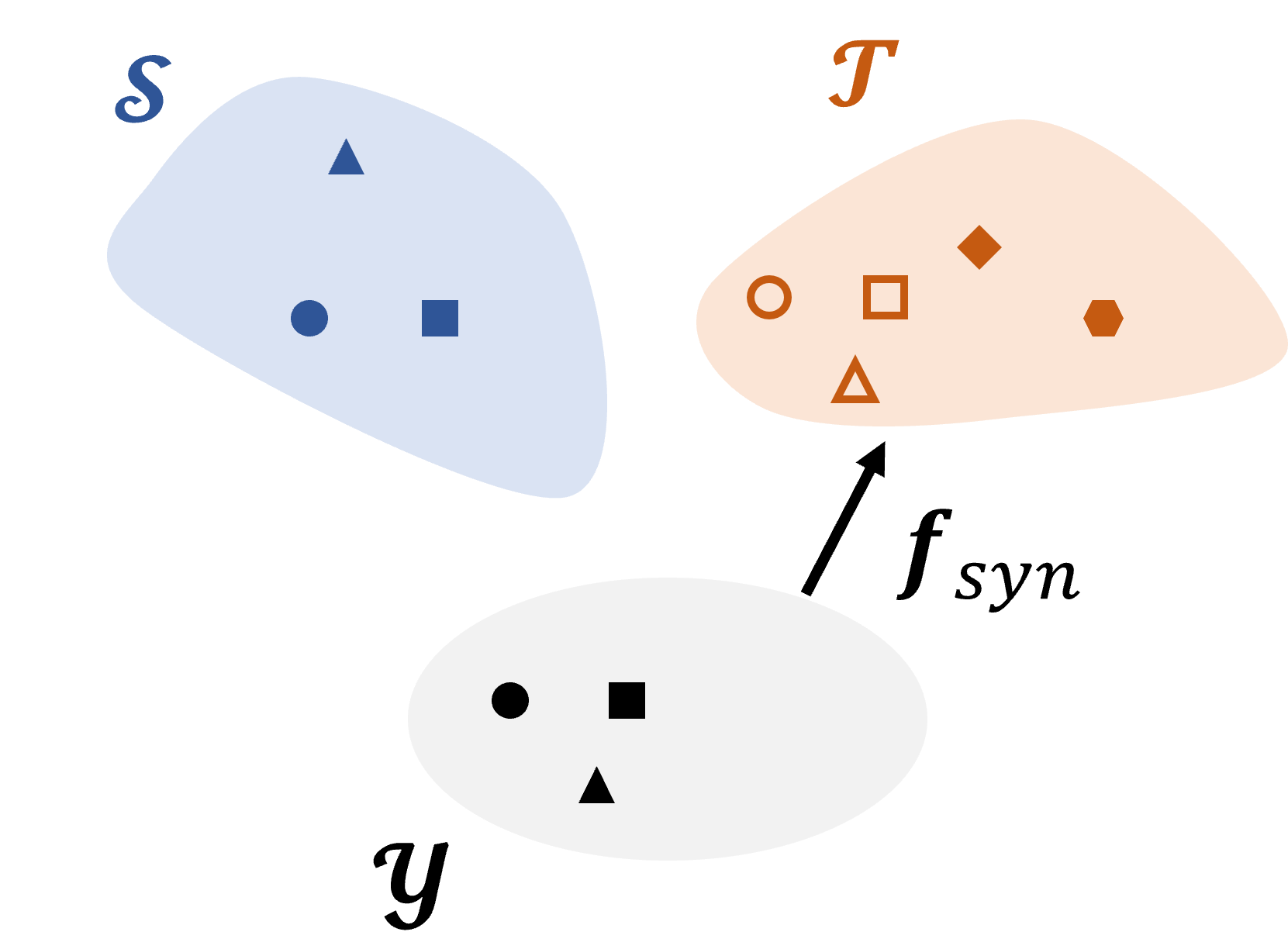}&
\includegraphics[width=.24\linewidth]{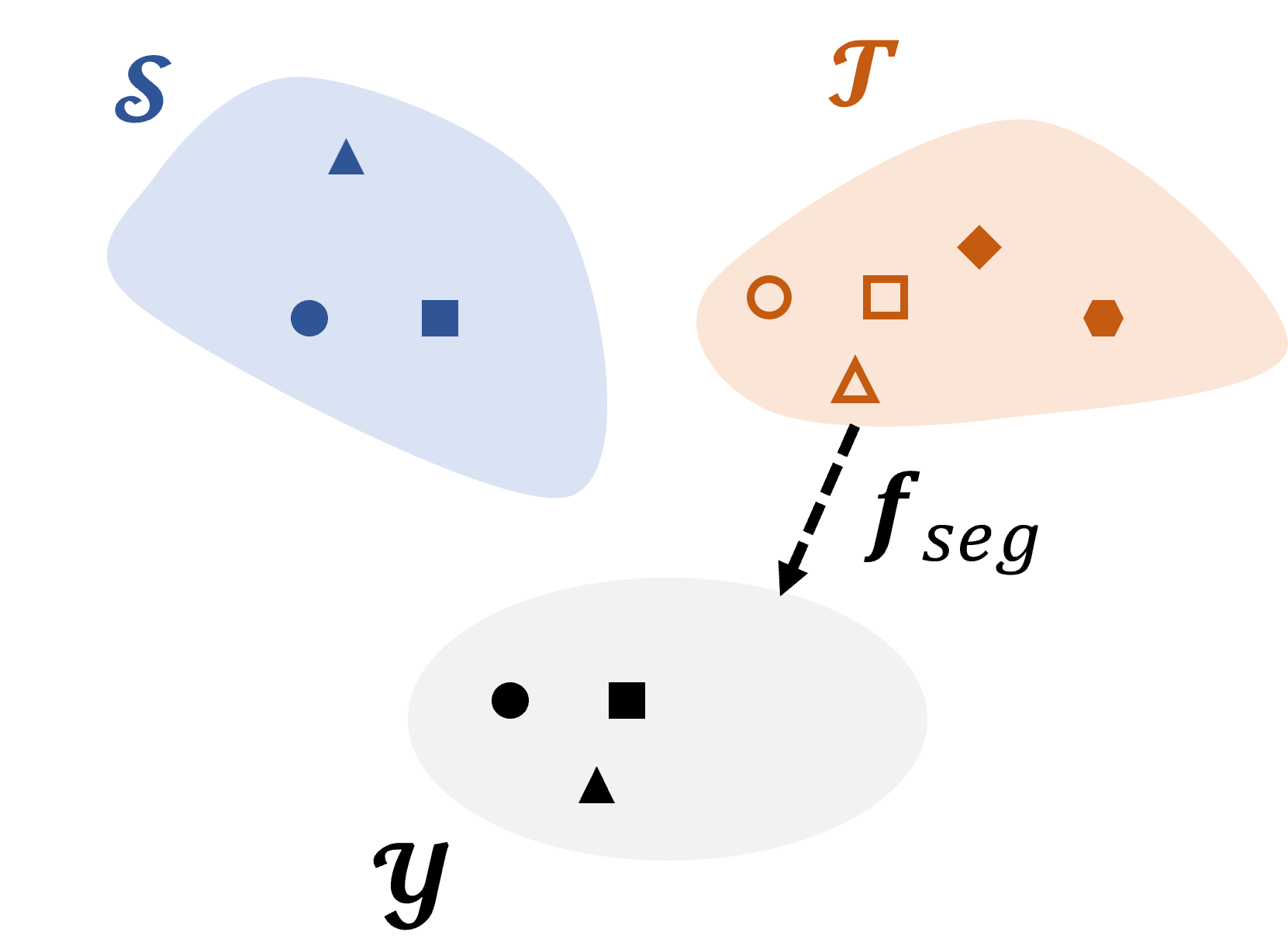}\\
(Step 1)&(Step 2)&(Step 3)&(Step 4)
\end{tabular}
\caption{\textbf{(Step 1)} Train the segmentation model $f_{seg}$ on labeled source domain and test on target domain images to create pseudo-labels. \textbf{(Step 2)} Train a semantic conditional diffusion model $f_{syn}$ with $\{\mathbf{x}^{\mathcal{T}}, \hat{\mathbf{y}}\}$. \textbf{(Step 3)} Inference the synthetic model to generate target domain samples corresponding to the real labels $\mathbf{y}$. \textbf{(Step 4)} Fine-tune the segmentation model $f_{seg}$ on the target domain with $\{\hat{\mathbf{x}}^{\mathcal{T}}, \mathbf{y}\}$. \textit{Dashed/solid lines}: model training/testing. Different \textit{marker shapes} represent distinct anatomies. \textit{Solid} shapes are real images and manual labels, \textit{outlines} are synthetic images and pseudo-labels.}
\label{fig:algorithm}
\end{figure}

\subsection{Assumption and notations}

In this study, we use multiple labeled FP datasets as the source domain denoted as $\{\mathbf{x}^\mathcal{S}_i, \mathbf{y}_i\}_{i=1}^{|\mathcal{S}|}$ where $|\mathcal{S}|$ represents the total number of paired data. The target domains are unlabeled OCT-A or FA images, denoted as $\{\mathbf{x}^\mathcal{T}_j\}_{j=1}^{|\mathcal{T}|}$. Since all three modalities are utilized to image the human retinal vasculature, we assume that there is no significant structural difference between underlying ground truth across different domains. In other words, \textbf{we ignore the potential distribution shift in binary vessel masks $\mathbf{y}$ from different datasets.}

\subsection{UDA via image synthesis}

Based upon the assumption above, we aim to develop a synthetic model $f_{syn}$ that can generate a realistic target domain image conditioned on a given binary mask, i.e., $f_{syn}:\mathbf{y} \rightarrow \mathbf{x}^{\mathcal{T}}$. The generated target image is denoted as $\hat{\mathbf{x}}^{\mathcal{T}}=f_{syn}(\mathbf{y})$. Note that we can use the existing binary vessel masks $\mathbf{y}$ in the labeled source domain. In this way, we can acquire paired data $\{\hat{\mathbf{x}}^{\mathcal{T}}, \mathbf{y}\}$ on any unlabeled unseen dataset and conduct supervised learning to get a domain adapted segmentation model. Since it has been proved that the diffusion probabilistic model is capable of superior synthesis performance over generative adversarial networks (GANs) \cite{dhariwal2021diffusion}, we propose a semantic conditional diffusion model for $f_{syn}$ so that the resultant image is paired with the label. However, in all the previous approaches \cite{yu2023diffusion,oh2023diffmix}, the $f_{syn}$ is trained with annotated data, which requires labels on target domains. In this work, we relax this constraint and demonstrate that $f_{syn}$ can be trained in weakly supervised condition in Sec.~\ref{Sec:weak supervision}. Fig.~\ref{fig:algorithm} illustrates our four steps to get a domain-specific segmentation model on any unseen dataset. 

\textbf{Step 1}: We train a segmentation model $f_{seg}$ with the labeled source domain data $\{\mathbf{x}^\mathcal{S}_i, \mathbf{y}_i\}_{i=1}^{|\mathcal{S}|}$. Specifically, we pre-process the green channel of the FP image with CLAHE \cite{reza2004realization} to enhance the contrast between the vessels and the background. Also, we invert the intensity so that the vessels are bright. Then, the model is tested on the target domain images $\{\mathbf{x}_j^{\mathcal{T}}\}_{j=1}^{|\mathcal{T}|}$ to create corresponding pseudo-labels $\{\mathbf{\hat{y}}_j\}_{j=1}^{|\mathcal{T}|}$. \textbf{Step 2}: With the pseudo-labels, we are able to train the conditional diffusion model $f_{syn}$ with $\{\mathbf{x}^\mathcal{T}_j, \hat{\mathbf{y}}_j\}_{j=1}^{|\mathcal{T}|}$ in a weakly supervised manner. In Sec.~\ref{Sec:weak supervision}, we show that the imperfect pseudo-labels $\hat{\mathbf{y}}$ and even noisier labels are sufficient for the model to learn about the conditional semantic information. \textbf{Step 3}: We input the source domain labels $\{\mathbf{y}_i\}_{i=1}^{|\mathcal{S}|}$ as semantic conditional information to the diffusion model $f_{syn}$ to generate synthetic target data $\hat{\mathbf{x}}_i^{\mathcal{T}}=f_{syn}(\mathbf{y}_i)$. \textbf{Step 4}: Fine-tune the segmentation model $f_{seg}$ on the synthetic paired data on target domain $\{\hat{\mathbf{x}}_i^{\mathcal{T}}, \mathbf{y}_i\}_{i=1}^{|\mathcal{T}|}$.

\subsection{Weak conditional diffusion}
\label{Sec:weak supervision}

\begin{figure}[t]
    \centering
    \includegraphics[width=.72\linewidth]{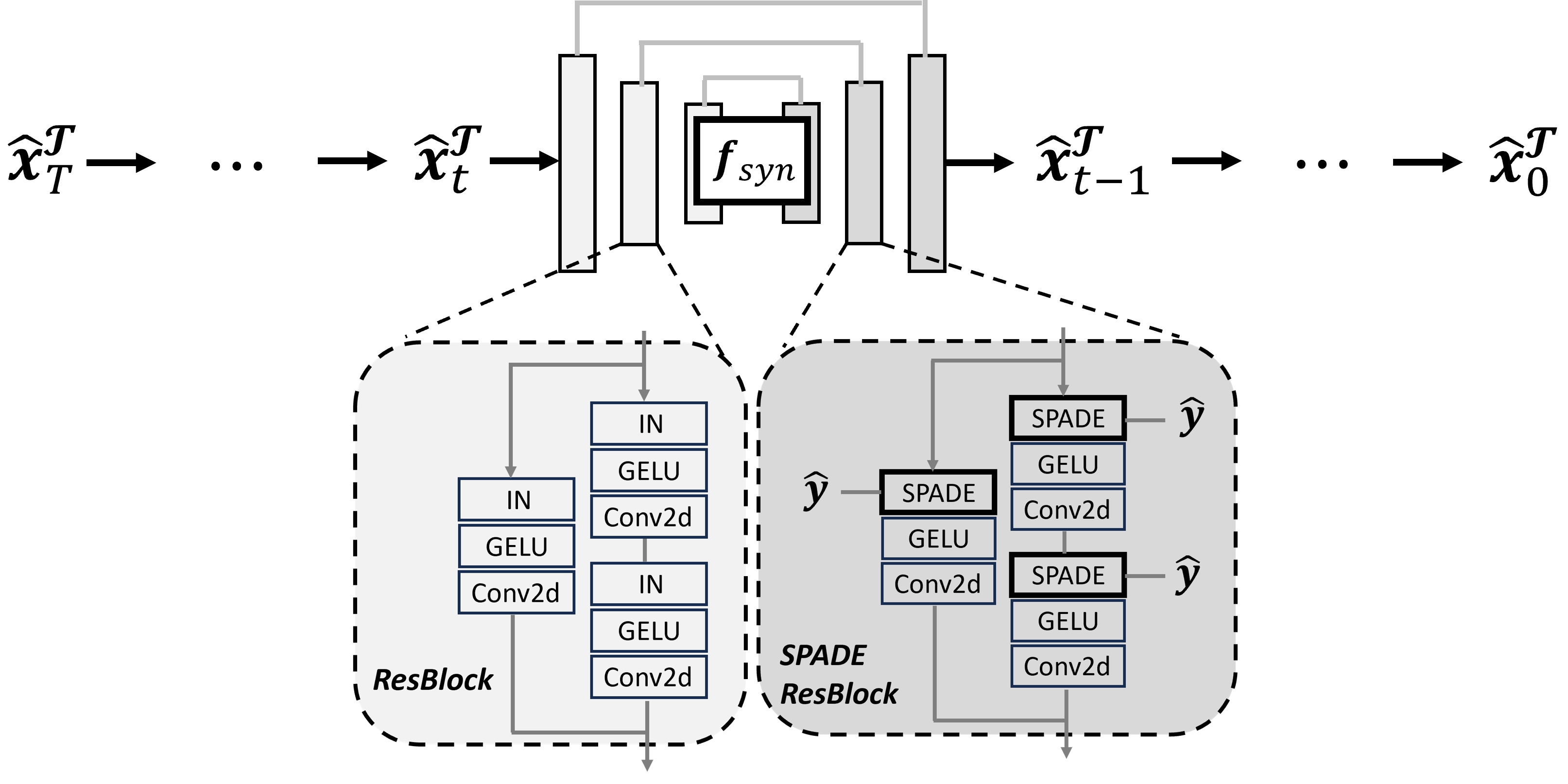}
    \caption{Weak conditional diffusion model. The semantic condition $\hat{\mathbf{y}}$ is added to the residual U-Net model by the spatial normalization block (SPADE)\cite{park2019semantic}}. 
    \label{fig:model}
\end{figure}

For the synthetic model $f_{syn}$, we implement the semantic conditional diffusion model illustrated in Fig.~\ref{fig:model} to generate $\hat{\mathbf{x}}^{\mathcal{T}}$. Instead of using the classifier-free guidance \cite{wang2022semantic}, we utilize a weak supervision with pseudo-label $\hat{\mathbf{y}}$. During training, Gaussian noise is added to the real target domain image $\mathbf{x}^{\mathcal{T}}$ with respect to a time step $t \in [1, T]$ in a fixed increasing variance schedule defined by $\{\beta_1, \cdots, \beta_T\}$, where $\beta_t \in (0, 1)$. Since the sum of Gaussians is still a Gaussian, the forward process is then defined by:
\begin{equation}
    \mathbf{x}^\mathcal{T}_{t} = \sqrt{\bar{\alpha}_t}\mathbf{x}_0^\mathcal{T} + \sqrt{1-\bar{\alpha}_t}\mathbf{\epsilon}, \quad \mathbf{\epsilon} \sim \mathcal{N}(0, \mathbf{I})
\end{equation}
where $\alpha_t=1-\beta_t$ and $\bar{\alpha}_t=\prod_{i=1}^t\alpha_t$. The synthetic model is trained to predict the noise $\mathbf{\epsilon}$ of each step $t$ in the reverse process, i.e., $\mathbf{\hat{\epsilon}}=\epsilon_{\theta}(\hat{\mathbf{x}}^{\mathcal{T}}_t, \hat{\mathbf{y}}, t)$, where $\hat{\mathbf{y}}=f_{seg}(\mathbf{x}^{\mathcal{T}})$ is the pseudo-label and $\theta$ represents the learnable parameters in the model. The reverse process step uses reparameterization sampling:
\begin{equation}
\label{eq:reverse}
    \hat{\mathbf{x}}_{t-1}^{\mathcal{T}} = \frac{1}{\sqrt{\alpha_t}}\left[\hat{\mathbf{x}}_t^{\mathcal{T}}-\frac{\beta_t}{\sqrt{1-\bar{\alpha}}_t}\epsilon_{\theta}(\hat{\mathbf{x}}^{\mathcal{T}}_t, \hat{\mathbf{y}}, t)\right]+\frac{\beta_t(1-\bar{\alpha}_{t-1})}{1-\bar{\alpha}_t}\delta
\end{equation}
where $\delta \sim \mathcal{N}(0, \mathbf{I})$. Note that the conditional image synthesis $f_{syn}(\mathbf{y})=\hat{\mathbf{x}}^{\mathcal{T}}$ is achieved by iteratively sampling $T$ times from $\hat{\mathbf{x}}_{T}^{\mathcal{T}}$ to $\hat{\mathbf{x}}_{0}^{\mathcal{T}}$ using Eq.~\ref{eq:reverse}.  

       

        

\begin{figure}[t]
    \centering
    \includegraphics[width=0.99\linewidth]{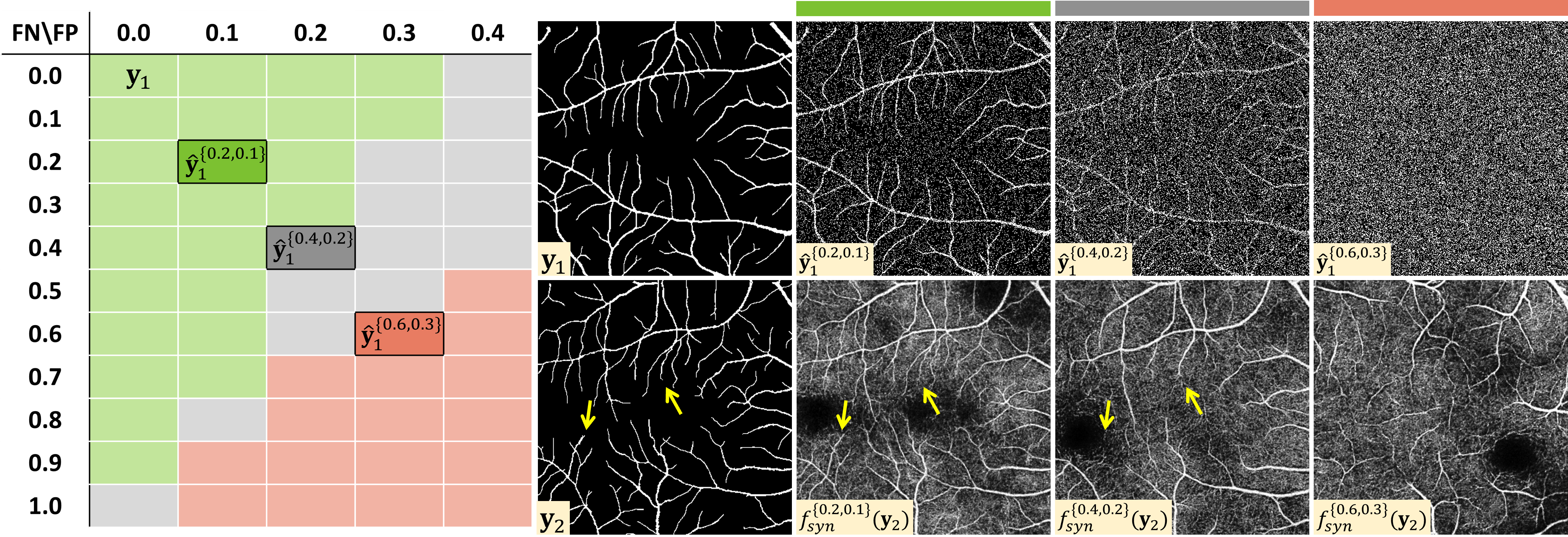}
    \caption{\textbf{Left}, performance of the semantic diffusion model trained with polluted label as semantic conditions. \textbf{Green:} cases where the generated images are visually well-correlated with the label during testing. \textbf{Gray:} borderline cases where some vessels are missed during testing. \textbf{Red:} cases where the model fails. \textbf{Right}, highlighted examples from each performance level (black-outlined cells from left panel). \textbf{Top row:} the increasingly polluted labels of $\mathbf{y}_1$ during training. \textbf{Bottom row:} a test label $\mathbf{y}_2$ and the sampled images $f_{syn}(\mathbf{y}_2)$ during testing. The yellow arrows highlight  missed vessels in $f^{\{0.4,0.2\}}_{syn}(\mathbf{y}_2)$. $f^{\{0.6,0.3\}}_{syn}(\mathbf{y}_2)$ does not correlate well with the input label $\mathbf{y}_2$.}
    \label{fig:polluted_label}
\end{figure}

In our experiments, we show that the semantic condition $\mathbf{y}$ does not necessarily need to be rigorously labeled by experts. Instead, pseudo-labels provided by $f_{seg}$ trained on source domain are sufficient to help the diffusion network learn about the correlation between the semantic mask and the image. To examine the robustness of the conditional diffusion model to imperfections on the labels used as condition, we create low-quality binary vessel masks by randomly adding increasing amounts of false positives (FP) and false negatives (FN) to the original labels $\mathbf{y}$ (Fig.~\ref{fig:polluted_label} right, top row). For each noise level, we train a weakly conditional diffusion model using the OCTA-500 (6M) as the target dataset. During inference, we use each model to generate an image conditioned on the original binary mask from the STARE source dataset. If the noise level of the label lies in the green region (Fig.~\ref{fig:polluted_label} left), the sampled image during inference keeps the desired anatomy and can be regarded as a good training pair. As the pseudo-label gets worse, in the gray region in Fig.~\ref{fig:polluted_label}, many small vessels are missed in the synthesized image (yellow arrows). For extreme noise (red region), the sampled data is not controlled by the semantic mask. In general, the proposed conditional model is robust with regard to noisy masks within a realistic range. In practice, the pseudo-labels are unlikely to be as bad as the last two cases in Fig.~\ref{fig:polluted_label}, so the $f_{syn}$ is able to generate realistic paired data on target domains.   

\section{Experiment}
\label{Sec:experiment}

\noindent
\normalfont{\textbf{Datasets.}} 
We use 7 public datasets (Table \ref{tab:dataset}). The source domain $\mathcal{S}$ contains 3 FP datasets: DRIVE, STARE and HRF. The target domain $\mathcal{T}$ contains 2 OCTA datasets, OCTA500 and ROSE, and 2 FA datasets, RECOVERY and PRIME. Since PRIME lacks manual labels, we use it for qualitative evaluation only.

\begin{table}[t]
\centering
\begin{tabular}{p{0.25\textwidth}>{\centering}
                p{0.15\textwidth}>{\centering}
                p{0.20\textwidth}>{\centering\arraybackslash}
                p{0.12\textwidth}>{\centering\arraybackslash}
                p{0.12\textwidth}}
\specialrule{.1em}{.05em}{.05em}
\scriptsize{\hspace{2em}\textbf{dataset}} & \scriptsize{\textbf{modality}} & \scriptsize{\textbf{resolution}} & \scriptsize{\textbf{number}} & \scriptsize{\textbf{domain}} \\
\hline
\scriptsize{DRIVE \cite{staal2004ridge}} & \scriptsize{fundus} & \scriptsize{$565\times 584$} & \scriptsize{20} & \scriptsize{$\mathcal{S}$}\\
\scriptsize{STARE \cite{hoover2000locating}} & \scriptsize{fundus} & \scriptsize{$700\times 605$} & \scriptsize{20} & \scriptsize{$\mathcal{S}$}\\
\scriptsize{HRF\cite{budai2013robust}} & \scriptsize{fundus} & \scriptsize{$3504\times 2336$} & \scriptsize{$45$} & \scriptsize{$\mathcal{S}$}\\
\rowcolor{Gray1}
\scriptsize{{ROSE} \cite{ma2020rose}} & \scriptsize{OCT-A} & \scriptsize{$304\times 304$} & \scriptsize{30} & \scriptsize{$\mathcal{T}$}\\
\rowcolor{Gray1}
\scriptsize{{OCTA-500(3M)} \cite{li2020image}} & \scriptsize{OCT-A} & \scriptsize{$400\times 400$} & \scriptsize{200} & \scriptsize{$\mathcal{T}$}\\
\rowcolor{Gray1}
\scriptsize{{OCTA-500(6M)} \cite{li2020image}} & \scriptsize{OCT-A} & \scriptsize{$400\times 400$} & \scriptsize{300} & \scriptsize{$\mathcal{T}$}\\
\rowcolor{Gray1}
\scriptsize{{RECOVERY-FA19} \cite{ding2020novel}} & \scriptsize{FA} & \scriptsize{$3900\times 3072$} & \scriptsize{8} & \scriptsize{$\mathcal{T}$}\\
\rowcolor{Gray1}
\scriptsize{{PRIME} \cite{ding2020weakly}} & \scriptsize{FA} & \scriptsize{$4000\times 4000$} & \scriptsize{$15$} & \scriptsize{$\mathcal{T}$}\\
\specialrule{.1em}{.05em}{.05em}
\end{tabular}
\caption{Datasets. Rows indicating the source domains have a white background while the target domains are shaded.}
\label{tab:dataset}
\end{table}

\noindent
\normalfont{\textbf{Implementation details.}}
The diffusion model $f_{syn}$ is a U-Net architecture with spatially adapted normalization blocks that extract semantic information from $\hat{\mathbf{y}}$ and a temporal sinusoidal encoding for time step $t$. For training, we set up the noise schedule of $T=300$ with a linearly increasing variance, $\beta_0=1\times 10^{-4}$ and $\beta_{T}=0.02$. The model is trained for 100 epochs with mean square (MSE) loss. The initial learning rate is $1\times 10^{-3}$ and will decay by $0.5$ for every 5 epochs.
The segmentation network $f_{seg}$ is a residual UNet. The fine-tuning takes 20 epochs with Dice loss and cross-entropy loss. The initial learning rate is $5\times 10^{-3}$ and will decay by $0.5$ for every 4 epochs.
Both models are trained and tested on an NVIDIA RTX 2080TI 11GB GPU. 

\section{Results}
\label{Sec:result}

\subsection{Quantitative evaluation}

\renewcommand{\arraystretch}{0.95}
\begin{table}[H]
    \centering
    \begin{tabular}{p{0.15\textwidth}>{\centering}
                    p{0.2\textwidth}>{\centering\arraybackslash}
                    p{0.2\textwidth}>{\centering\arraybackslash}
                    p{0.2\textwidth}>{\centering\arraybackslash}
                    p{0.2\textwidth}}
        \specialrule{.1em}{.05em}{.05em}
        
        \hspace{0.1em} {\scriptsize Method} & {\scriptsize OCTA 500(3M)} & {\scriptsize OCTA 500(6M)} & {\scriptsize ROSE} & {\hspace{-0.2em}\scriptsize RECOVERY}\\
         
        \specialrule{.1em}{.05em}{.05em}
        \scriptsize \hspace{0.08em} \textit{baseline} & $\scalemath{0.9}{65.97 \pm 7.56}$ & $\scalemath{0.9}{73.16 \pm 4.54}$ & $\scalemath{0.9}{67.41 \pm 3.35}$ & $\scalemath{0.9}{60.98 \pm 4.20}$ \\
        \hline
        \scriptsize \hspace{0.3em} DANN~\cite{javanmardi2018domain} & $\scalemath{0.9}{69.27 \pm 9.81}$ & $\scalemath{0.9}{80.93 \pm 3.31}$ & $\scalemath{0.9}{70.37 \pm 8.21}$ & $\scalemath{0.9}{66.33 \pm 3.77}$ \\
        \scriptsize CycleGAN~\cite{palladino2020unsupervised}  & $\scalemath{0.9}{31.97 \pm 3.90}$ & $\scalemath{0.9}{37.47 \pm 3.13}$ & $\scalemath{0.9}{16.05 \pm 2.31}$ & $\scalemath{0.9}{22.95 \pm 2.41}$ \\
        \scriptsize \hspace{0.65em} CUT~\cite{park2020contrastive} & $\scalemath{0.9}{28.44 \pm 1.88}$ & $\scalemath{0.9}{39.82 \pm 1.46}$ & $\scalemath{0.9}{14.87 \pm 1.40}$ & $\scalemath{0.9}{28.33 \pm 1.02}$ \\
        \scriptsize \hspace{0.25em} SynSeg~\cite{huo2018synseg}  & $\scalemath{0.9}{44.15 \pm 3.86}$ & $\scalemath{0.9}{47.99 \pm 3.29}$ & $\scalemath{0.9}{42.42 \pm 3.09}$ & $\scalemath{0.9}{36.68 \pm 2.57}$ \\
        \scriptsize \hspace{0.05em} Proposed & $\scalemath{0.9}{\bf{72.73 \pm 5.34}}^{\dagger}$ & $\scalemath{0.9}{\bf{81.94 \pm 2.76}}^{\dagger}$ & $\scalemath{0.9}{\bf{76.15 \pm 2.55}}^{\dagger}$ & $\scalemath{0.9}{\bf{71.38 \pm 2.72}}^{\dagger}$ \\
        \hline
        \scriptsize \hspace{0.5em} \textit{oracle} & $\scalemath{0.9}{87.61 \pm 2.13}$ & $\scalemath{0.9}{84.78 \pm 2.79}$ & $\scalemath{0.9}{78.74 \pm 2.47}$ & $\scalemath{0.9}{77.13 \pm 2.72}$ \\
        \specialrule{.1em}{.05em}{.05em}
    \end{tabular}
    \caption{Dice scores (\%) for testing on target domains. Boldface: best result, underline: second best result. $^\dagger: \text{p-value} \ll 0.05$ in paired t-test compared to the baseline.}
    \label{tab:result}
\end{table}

In Table \ref{tab:result}, we evaluate the UDA  segmentation performance by the Dice score. The baseline model tests $f_{seg}$ on target images after Step 1. The oracle model is acquired by training directly on the target domain. Since the proposed AdaptDiff can generate realistic target domain images that closely match  the label, its adaptation performance is the best in all target datasets. For OCTA500 (6M) and ROSE, our results are very close to the oracle model which is directly trained on the target domain. DANN has a reasonable performance above the baseline. We note that CycleGAN,  CUT and SynSeg have even lower performance than the baseline. To understand this behavior, we next look at qualitative results.

\setlength{\tabcolsep}{1pt}
\begin{figure}[t]
    \centering
    \begin{tabular}{cccccc}
         $\mathbf{x}^\mathcal{S}$ & $\mathbf{y}$ & {\scriptsize \textbf{CycleGAN}} & 
         {\scriptsize \textbf{CUT}} &{\scriptsize \textbf{SynSeg}} & {\scriptsize \textbf{AdaptDiff}} \\

         \includegraphics[width=0.161\linewidth]{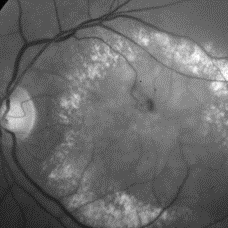} &
         \includegraphics[width=0.161\linewidth]{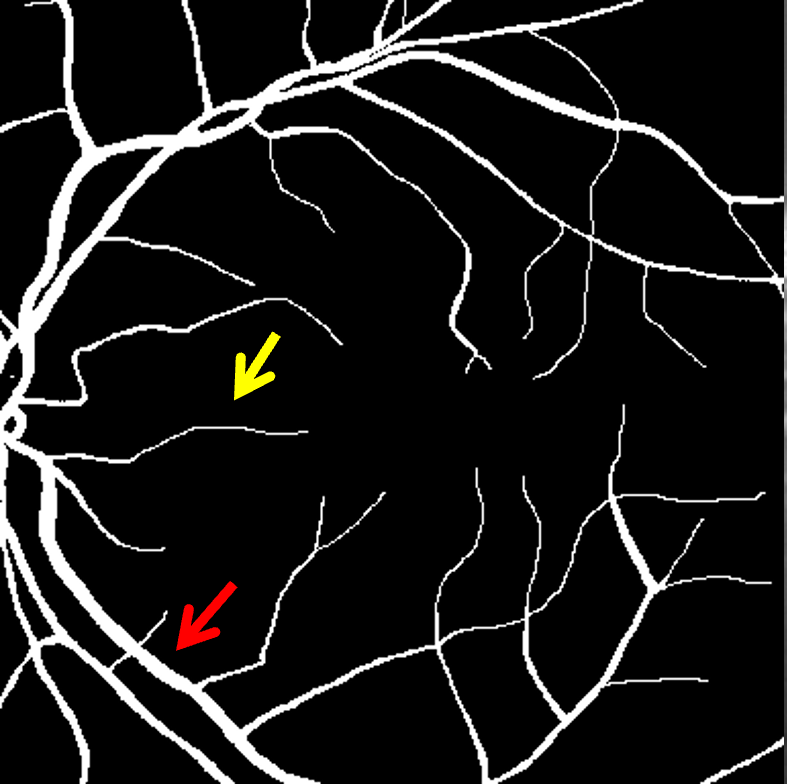} & 
         \includegraphics[width=0.161\linewidth]{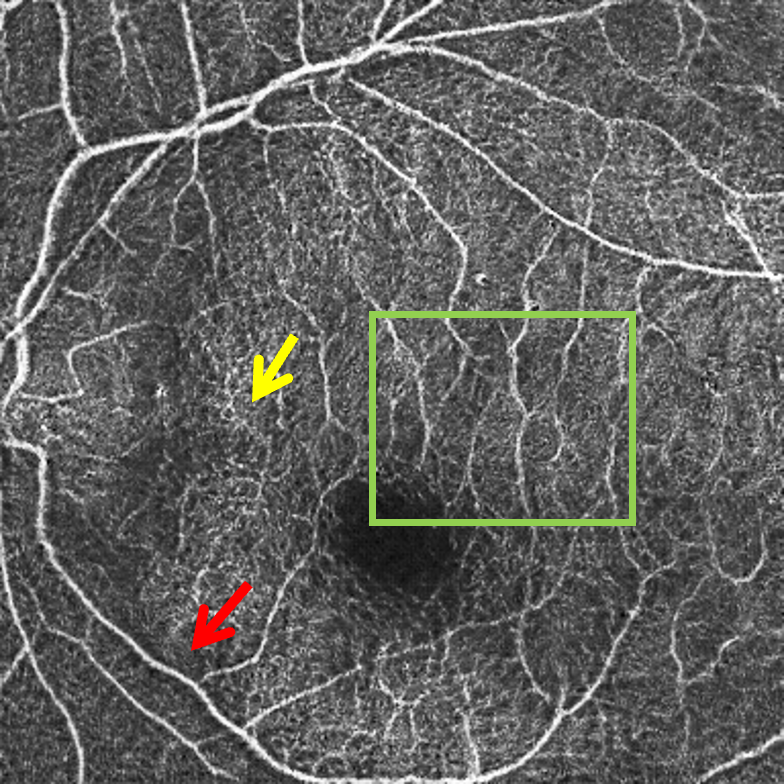} &
         \includegraphics[width=0.161\linewidth]{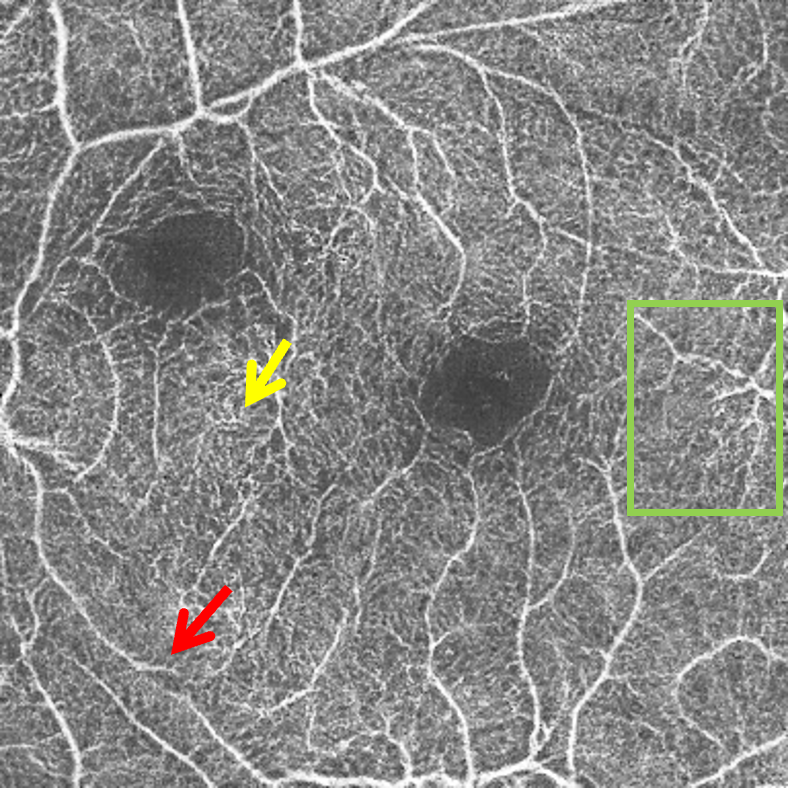} &
         \includegraphics[width=0.161\linewidth]{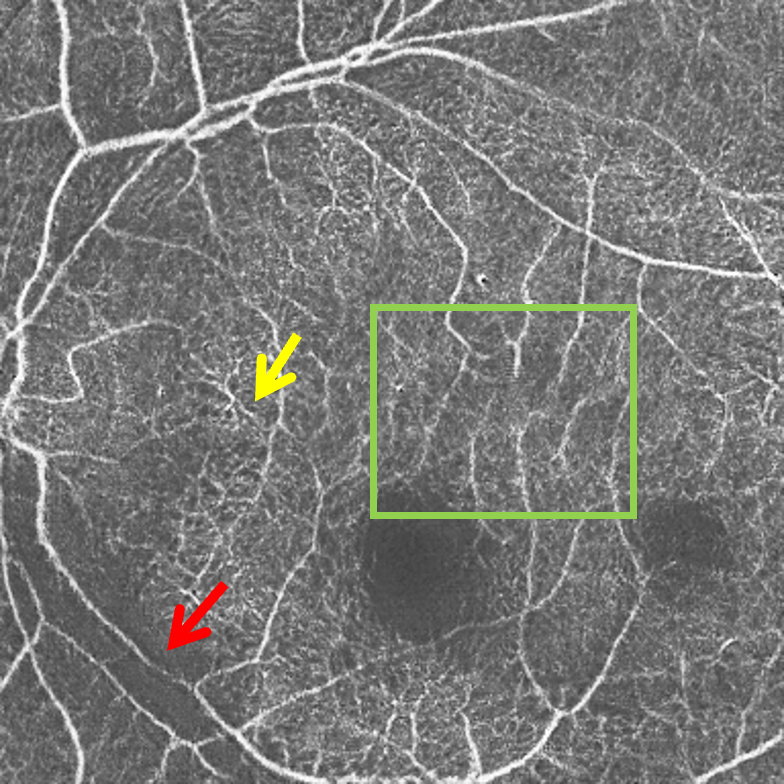} &
         \includegraphics[width=0.161\linewidth]{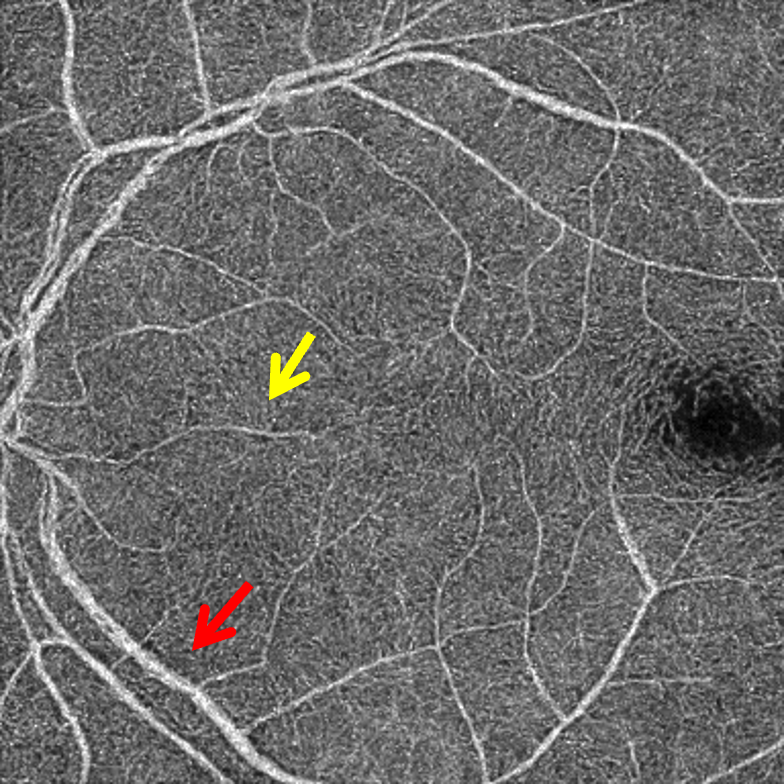}\\

         \includegraphics[width=0.161\linewidth]{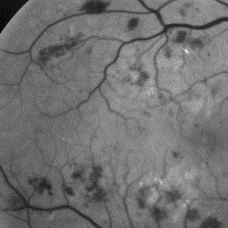} &
         \includegraphics[width=0.161\linewidth]{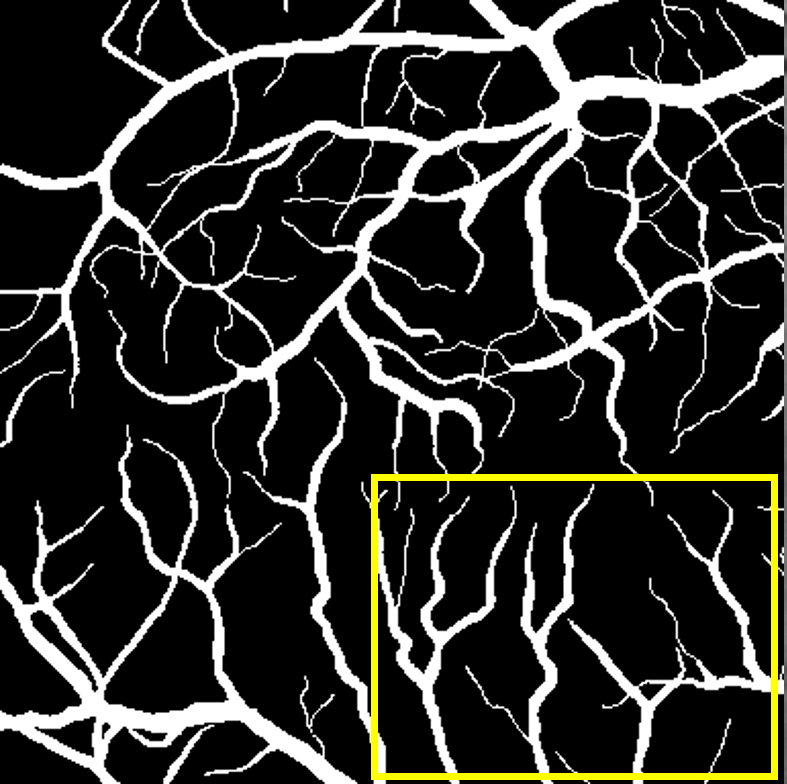} &
         \includegraphics[width=0.161\linewidth]{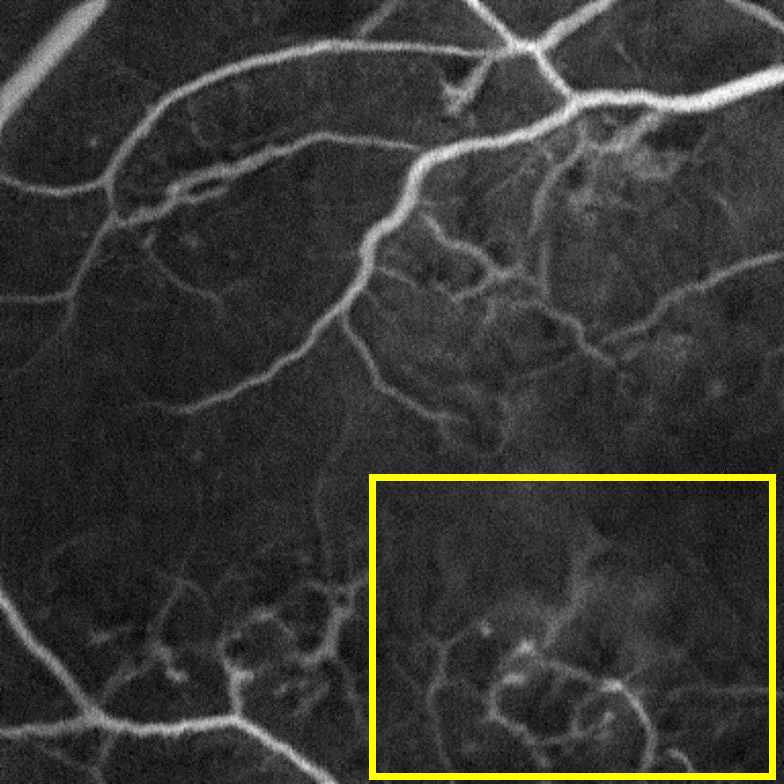} &
         \includegraphics[width=0.161\linewidth]{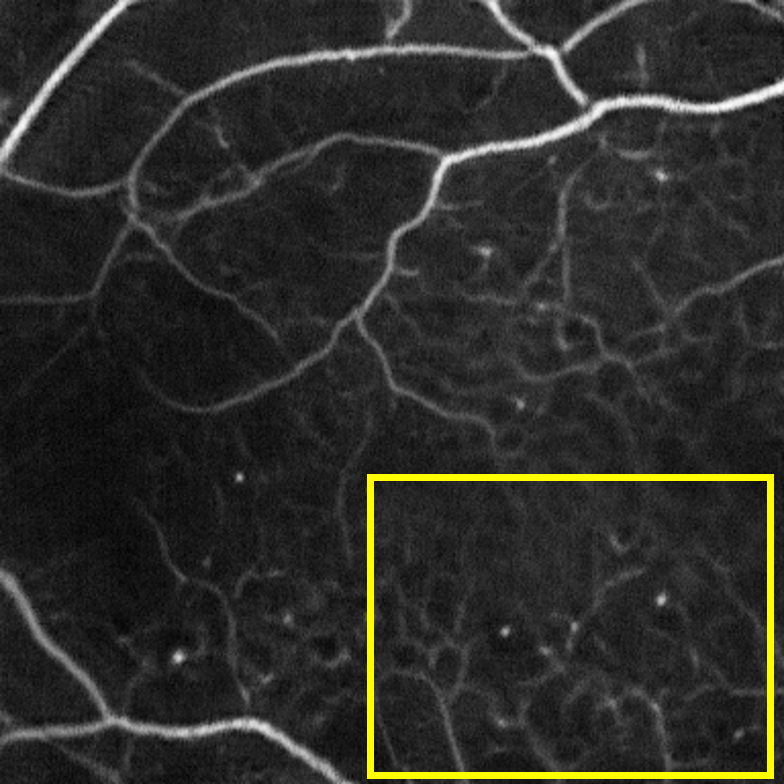} &
         \includegraphics[width=0.161\linewidth]{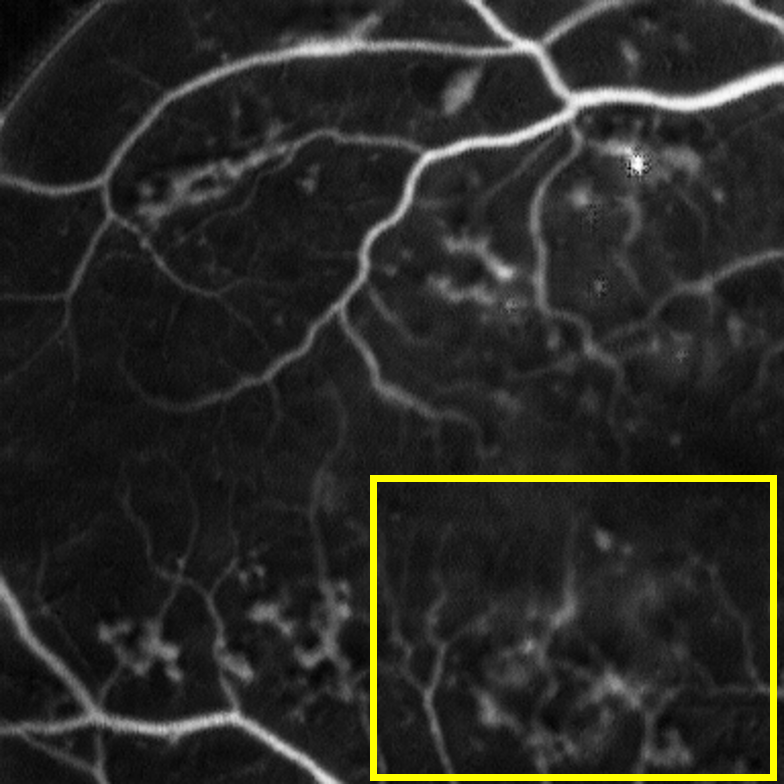} &
         \includegraphics[width=0.161\linewidth]{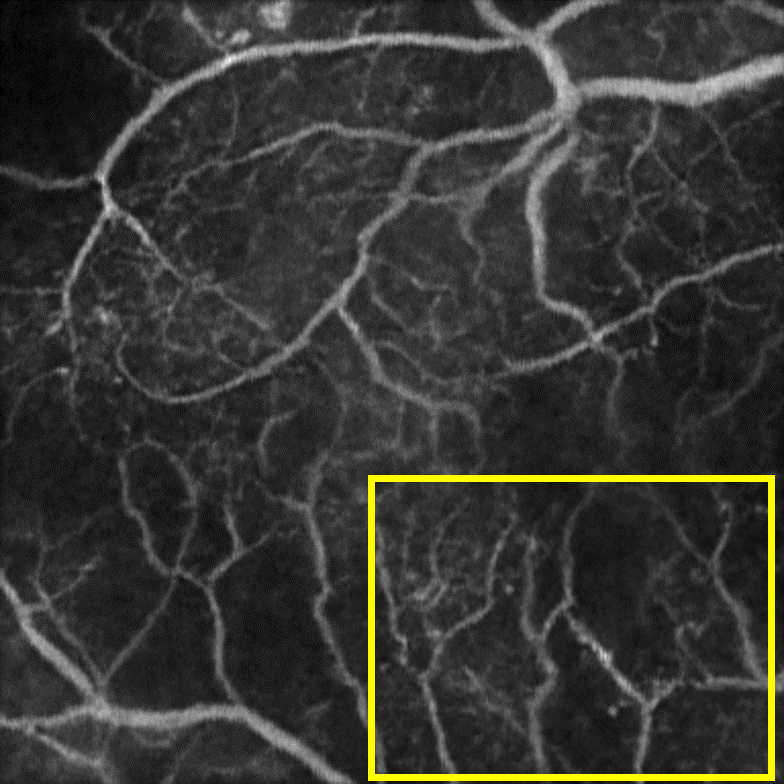}\\
         
    \end{tabular}
    \caption{Qualitative results. \textbf{Top row}, generated OCT-A images. Yellow arrows: preservation of a thin vessel. Red arrows: accuracy of vessel thickness. Green box:  hallucinated vessels. \textbf{Bottom row}, generated FA images. Yellow boxes highlight that CycleGAN, CUT and SynSeg are not able to capture the vessels due to the poor contrast in $\mathbf{x}^{\mathcal{S}}$. AdaptDiff-synthesized images have the correct anatomy in each of these cases.}
    \label{fig:qualitative}
\end{figure}

\subsection{Qualitative evaluation}
In Fig.~\ref{fig:qualitative} we compare the quality of the synthesized images on different target domains generated by the four image alignment approaches. We observe that the fake images generated by CycleGAN, CUT and SynSeg are not well correlated with the label. We highlight the three types of mismatch with markers in different color. Yellow arrows and boxes show missing vessels. This is usually caused by the poor contrast in the source domain images. Red arrows are pointing to an example of inaccurate vessel thickness in the generated image. Such inconsistency severely deteriorates the segmentation performance. The green boxes highlight a hallucination in the region without strong semantic control. The competing methods tend to fill the region with fake vessels.  In contrast, AdaptDiff successfully synthesizes images that match the provided vessel labels. 

Unlike brain CT-MR translation, retinal images have a wide variability of vessel anatomy that falls within the image field of view. Consequently, it is hard to learn the mapping function between these modalities with structural features preserved. Therefore, mapping from the label space to the image domain can be a more effective solution than CycleGAN and similar methods.

\section{Conclusion}
\label{Sec:conclusion}

We proposed  AdaptDiff, a diffusion-based solution to the cross-modality UDA problem. We empirically showed that the conditional semantic diffusion model can be trained in a weakly supervised manner. Thus, for each unseen domain, we can generate paired data to fine-tune the segmentation model to significantly alleviate the effects of domain shift. Accounting for any domain shifts in the label space remains as future work.

\noindent
\normalfont{\textbf{Acknowledgements.}} This work is supported, in part, by the NIH grant R01-EY033969.

\clearpage
%
\bibliographystyle{splncs04}
\bibliography{refs.bib}

\end{document}